\documentclass[journal]{IEEEtran}
\usepackage{amsmath,amsfonts}
\usepackage{algorithmic}
\usepackage{algorithm}
\usepackage{array}
\usepackage[caption=false,font=normalsize,labelfont=sf,textfont=sf]{subfig}
\usepackage{textcomp}
\usepackage{stfloats}
\usepackage{url}
\usepackage{verbatim}
\usepackage{graphicx}
\usepackage{balance}
\usepackage[colorlinks=true,linkcolor=black,citecolor=black,urlcolor=blue]{hyperref}

\usepackage{tabularx}
\newcolumntype{Y}{>{\centering\arraybackslash}X}

\usepackage{cite}
\newcommand{\bs}[1]{\boldsymbol{#1}}  
\newcommand{\ts}[1]{\text{#1}} 

\renewcommand{\baselinestretch}{0.89}

\begin{document}

\title{\fontsize{24pt}{23pt}\selectfont An Autonomous Subgram SMA-Based Swimmer\\

\thanks{The research presented in this article was partially funded by the Washington State University (WSU) Foundation and the Palouse Club through a Cougar Cage Award to \mbox{N.\,O.\,P.-A.}; and, the Koerner Family Foundation (KFF) through a graduate fellowship to \mbox{C.\,K.\,T.}. Additional funding was provided by the WSU Voiland College of Engineering and Architecture through a \mbox{start-up} fund to \mbox{N.\,O.\,P.-A.}.} %
\thanks{The authors are with the School of Mechanical and Materials Engineering, Washington State University (WSU), Pullman,\,WA\,99164,\,USA. Corresponding authors' \mbox{e-mail:}
{\tt conor.trygstad@wsu.edu}~(C.\,K.\,T.);
{\tt n.perezarancibia@wsu.edu} (N.\,O.\,P.-A.).}%
}
\author{Conor K. Trygstad, Francisco M. F. R. Gon\c{c}alves, and N\'estor O. P\'erez-Arancibia}

%



\maketitle

\begin{abstract}
We present the Swima, a bioinspired \mbox{$\bs{900}$-mg} swimmer propelled by two \mbox{$\bs{10}$-mg} \textit{high-work-density} (HWD) actuators driven by \textit{shape-memory \mbox{alloy}} (SMA) wires. We integrated onboard power and computation by using a \mbox{custom-built} \textit{printed circuit board}~(PCB) and an \mbox{$\bs{11}$-mAh} \mbox{$\bs{3.7}$-V} \mbox{$\bs{507}$-mg} \mbox{single-cell} \textit{\mbox{lithium-ion}} (\mbox{Li-Ion}) battery, which in conjunction enable autonomous swimming in excess of \mbox{$\bs{18}\,\ts{min}$}. The \mbox{Swima\hspace{-0.3ex}} can swim at speeds of up to \mbox{$\bs{22.4}$\,mm/s}~\mbox{($\bs{0.56}$\,Bl/s)}, achieves turning rates of up to \mbox{$\bs{14}$}\textdegree/s, and can follow $\bs{0}$-degree heading reference trajectories with \emph{root mean square}~(RMS) values of tracking errors of about $\bs{6.5}$\textdegree~across multiple tests. This robot is the first subgram microswimmer with onboard power, actuation, and computation developed to date. 
\end{abstract}


\section{Introduction}
\label{SECTION01}
\IEEEPARstart{T}{he} development of triphibious swarms of autonomous insect-scale robots has the potential to enhance humans' technological capabilities in several fields such as disaster response, infrastructure inspection, and pollution detection\mbox{\cite{ChitikenaH2023, AitkenJM2021, UrsoM2023}}. However, the majority of the microrobots reported to date either function physically tethered to power sources\mbox{\cite{ BenaRM2023, WuY2019, ZhouW2020, BenaRM2021, TrygstadCK2023, TrygstadCK2024, TrygstadCK2025, BlankenshipEK2024, RenZ2022, LiK2023, ChenY2017}} or rely on bulky, infrastructure-intensive external magnetic actuation systems that are impractical for use in unstructured environments\mbox{\cite{XuT2013, ZhangL2010, LiuW2010}}. While the creation of novel actuation and \mbox{energy-storage} technologies has contributed to the realization of new \mbox{mm-to-cm--scale} autonomous terrestrial vehicles\mbox{\cite{GoldbergB2018, JohnsonK2023, JiX2019, YangX2020}}, these technologies have not yet been leveraged to develop \mbox{insect-sized} \textit{\mbox{autonomous} underwater vehicles} (AUVs). On the other hand, a significant number of \mbox{cm-to-dm--scale} underwater vehicles showing great promise for deployment in unstructured fields have been produced in recent years\mbox{\cite{SpinoP2024, BerlingerF20210-Sci, BerlingerF2021-Bio, LiZ2024, GiorgioFS2013, CenL2013, WangT2023, VillanuevaA2011, LiuT2025, TangY2017}}; however, all these platforms use actuation methods and power sources that are not easily scalable to \mbox{mm-to-cm} sizes. 

\begin{figure}[t!]
\vspace{0.75ex}
\begin{center}
\includegraphics[width=\linewidth]{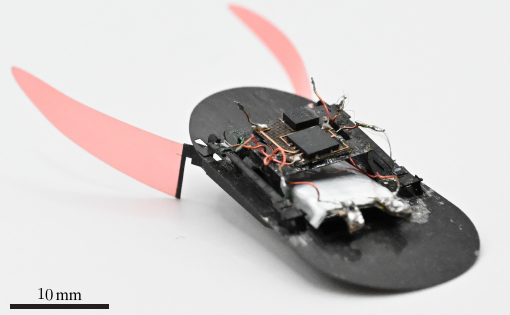}
\end{center}
\vspace{-2ex}
\caption{\textbf{An autonomous insect-scale surface swimmer.} The \mbox{Swima}, a \mbox{$900$-mg} \mbox{autonomous} surface swimmer driven by two \mbox{$10$-mg} \mbox{SMA-based} \mbox{actuators}. \label{FIG01}}
\vspace{-2.0ex}
\end{figure}

To close this technological gap, we present \mbox{the~Swima} (see Fig.\,\ref{FIG01}), a bioinspired \mbox{$900$-mg} swimmer propelled by two \mbox{$10$-mg} \textit{high-work-density} (HWD) actuators driven by \textit{shape-memory \mbox{alloy}} (SMA) wires. This robot is autonomous from both the power and control perspectives and is capable of swimming at speeds of $22.4$\,mm/s~($0.56$\,Bl/s) while consuming an average power of about $87\,\ts{mW}$, allowing it to operate continuously for up to $18\,\ts{min}$ on a single charge of an $14$-mAh single-cell \emph{lithium-ion}~(Li-Ion) battery. Additionally, onboard computation and sensing are embedded into the platform with a \emph{microcontroller unit}~(MCU) and \emph{inertial-measurement unit}~(IMU), thus allowing the Swima to regulate its heading during feedback-controlled swimming with \emph{root mean square}~(RMS) values of tracking errors of about \mbox{$6.5$\textdegree} across multiple tests. When preprogrammed to follow \mbox{$90$-degree} left and right-turn maneuvers, the Swima is capable of turning rates of up to \mbox{$14$\textdegree/s}. To the best of our knowledge, the Swima is the first subgram autonomous surface swimmer developed to date. 

While SMA-based actuators are known for their HWD, typically, researchers avoid using SMA-based actuators due to their low electrical-to-mechanical energy conversion efficiency, which is in the range of $1~\ts{to}~6\,\%$~\cite{ZiolkowskiA1993, JaniJM2014}. However, the actuators of the Swima require low excitation voltages ($1~\ts{to}~4$\,V) and therefore can be driven with $8$-mg \emph{\mbox{metal-oxide-semiconductor} \mbox{field-effect} transistors}~(MOSFETs) that have transmission efficiencies on the order of $99\,\%$. Piezoelectric and dielectric-based actuators require power electronics that, when created at the mm-scale, have poor transmission efficiencies. For example, the piezoelectric boost circuit used in \cite{KarpelsonM2012} has a mass of $20\,\ts{mg}$ and an efficiency on the order of $30\,\%$ and the dielectric boost circuit used in \cite{RenZ2023} has a mass of $127\,\ts{mg}$ and an efficiency of $10\,\%$. Moreover, even when these dielectric boost circuits are made at a larger scale, such as the \textit{printed circuit board}~(PCB) used in \cite{HartmanF2025}, which has a mass on the order of $5$\,g and an overall voltage amplification efficiency on the order of $10$\,\%. Other types of \mbox{low-voltage} actuation technologies are electromagnetic-based, such as the ones used in \cite{Liu2024BHMBot, Xu2025Muscle}, and ones based on \emph{\mbox{ionic-polymer} metal composites}~\mbox{(IPMC)\cite{GuoS2003,HubbardJJ2014,KimB2005, ChenA2010,ChenZ2011,ChenZ2012}}. While both require low voltages, when scaled down to the mm-scale, the transmission efficiency of electromagnetic motors drops drastically due to increased resistive heating. For example, the electromagnetic motor in \cite{Xu2025Muscle} has an \mbox{electrical-to-mechanical} energy conversion efficiency on the order of $0.4~\ts{to}~2.5\,\%$, which is drastically lower than SMAs. In the case of \mbox{IPMC-based} actuators, these devices exhibit significantly lower work densities than those of \mbox{SMA-based} \mbox{actuators\cite{HeQ2011}}. This fact becomes evident when the relative speeds of \mbox{IPMC-driven} swimming \mbox{platforms\cite{GuoS2003,HubbardJJ2014,KimB2005, ChenA2010,ChenZ2011,ChenZ2012}} are compared to those achieved by the Swima. Overall, from the empirical evidence presented here, we believe that in the case of insect-scale AUVs, SMA-based actuation is an advantageous choice.

For transparency and clarity, we briefly state the technological contributions presented in this article relative to our previous work. Here we build upon the preliminary results presented in \cite{LongwellCR2024} to include additional empirical evidence of power and control autonomy of the Swima platform. Namely, we present a control strategy for robots of this type, present feedback-controlled swimming experiments, and assess the battery lifetime while driving the robot.

The rest of the paper is organized as follows. Section\,\ref{SECTION02} discusses the design and fabrication of the Swima. Section\,\ref{SECTION03} presents and discusses the results of autonomous swimming experiments performed with the Swima. Last, Section\,\ref{SECTION04} concludes the presented results and discusses possible future research directions.

\section{Design and Fabrication}
\label{SECTION02}

\begin{figure*}[t!]
\begin{center}
\includegraphics[width=\textwidth]{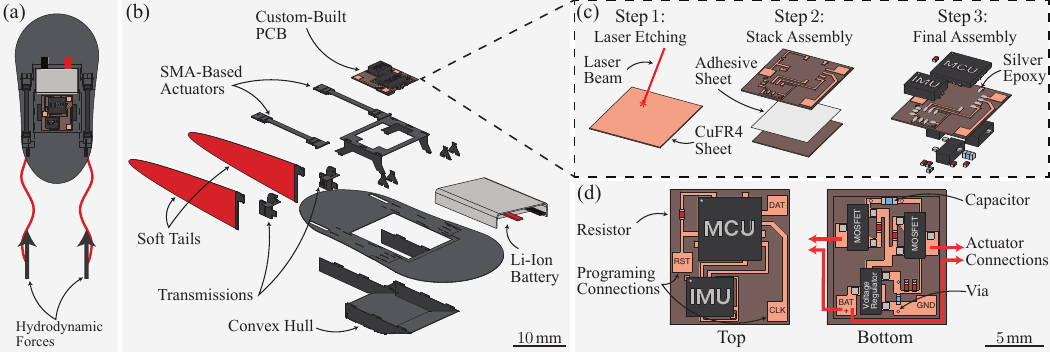}
\end{center}
\vspace{-2ex}
\caption{\textbf{Design and fabrication of the Swima.} \textbf{(a)}~Swimming mode of the Swima\hspace{-0.4ex}. The two \mbox{SMA-based} microactuators undulate the tails of the swimmer to produce the hydrodynamic forces and differentials required for forward propulsion and turning. \textbf{(b)}~Exploded view of the Swima's design. This robot is composed of six main types of components: (i)~two \mbox{SMA-based} microactuators with passive hinges installed at their ends; (ii)~two soft propulsors that propel the microswimmer forward; (iii)~two transmissions that amplify the output displacements generated by the robot's actuators into the two significantly larger stroke angles required to excite the two soft propulsors of the swimmer; (iv)~a convex hull made of carbon fiber that provides a buoyancy force to support the microswimmer on water; (v)~an \mbox{11-mAh} \mbox{$3.7$-V} \mbox{$507$-mg} \mbox{single-cell} \mbox{Li-Ion} battery (Powerstream~\mbox{GM$301014$H}); and, (vi)~the \mbox{custom-built} PCB. \textbf{(c)}~Fabrication process of the robot's \mbox{custom-designed} PCB. \mbox{In Step\,$1$}, sheets of \mbox{CuFR4} are etched using laser rasterization in order to remove areas of the \mbox{Cu-coating} and thus create the patterns designed for each side of the PCB. \mbox{In Step\,$2$}, the two sides of the PCB are \mbox{pin-aligned} and adhered together with a sheet of Pyralux adhesive by applying pressure and heat inside a curing oven. \mbox{In Step\,$3$}, the \mbox{off-the-shelf} elements composing the interconnected power, actuation, and computation circuits are installed on the two sides of the PCB using conductive silver epoxy (\mbox{MG-Chemicals~$8331$D}) and cured inside an oven. \textbf{(d)}~Top and bottom of the robot's PCB. The top side of the PCB supports an MCU for computation and an IMU for sensing. The bottom side of the PCB supports two \mbox{$\ts{N}$-channel} \mbox{MOSFETs} and a voltage regulator. The two sides of the PCB are connected through \textit{vias}. Decoupling capacitors and \mbox{pull-up/down} resistors are necessary to ensure the functionality of the circuits.} \label{FIG02}
\vspace{-2.0ex}
\end{figure*}

A schematic that explains the basic swimming mode of the \mbox{Swima\hspace{-0.3ex}~is} shown in \mbox{Fig.\,\ref{FIG02}(a)}. When the two soft tails of the propulsors are flapped, the resulting undulations produce, through \textit{fluid--structure interaction} (FSI), the hydrodynamic forces that propel the swimmer forward \cite{TrygstadCK2025ICAR}. By independently modulating the two \textit{\mbox{pulse-width} modulation} (PWM) signals that excite the two propulsors of the swimmer, we can vary the magnitude and direction of the total thrust acting on the system. In this manner, \textit{two-dimensional} (2D) controllability is achieved, as empirically demonstrated through the \mbox{feedback-controlled} swimming tests discussed in Section\,\ref{SECTION03B}.

An exploded view of the Swima design is shown in \mbox{Fig.\,\ref{FIG02}}(b). This robot consists of six main types of components: \mbox{(i)~two} \mbox{SMA-based} actuators that drive the propulsors of the swimmer; \mbox{(ii)~two} soft tails that generate the hydrodynamic forces required for locomotion through FSI; \mbox{(iii)~two} planar \mbox{four-bar} linkage transmissions that convert the displacement outputs of the actuators into large stroke angles that undulate the two soft tails of the swimmer; \mbox{(iv)~a} convex hull made of carbon fiber that provides a container to store the system's battery and generates a buoyancy force that works in parallel with the surface tension of water to maintain the swimmer afloat; \mbox{(v)~an} \mbox{11-mAh} \mbox{$3.7$-V} \mbox{$507$-mg} \mbox{single-cell} \mbox{Li-Ion} battery (Powerstream~\mbox{GM$301014$H}) that enables the swimmer to operate continuously and autonomously for up to $18\,\ts{min}$; and, \mbox{(vi)~the} \mbox{custom-built} PCB. The main element that enables tetherless \mbox{SMA-based} actuation for this swimmer is the PCB depicted in~\mbox{Figs.\,\ref{FIG02}(c)~and~(d)}. As seen in \mbox{Fig.\,\ref{FIG02}(c)}, the PCB is fabricated in three steps. In \mbox{Step\,1}, sheets of \textit{\mbox{copper-clad}~FR4}~(CuFR4) are laser etched to remove areas of the \mbox{Cu-coating} and thus create the patterns designed for each side of the PCB. In \mbox{Step\,2}, the two sides of the PCB are \mbox{pin-aligned} and adhered together with a sheet of Pyralux adhesive by applying pressure and heat inside a curing oven. In \mbox{Step\,3}, the \mbox{off-the-shelf} elements composing the interconnected power, actuation, and computation circuits are installed on the two sides of the PCB using conductive silver epoxy (\mbox{MG-Chemicals~$8331$D}) and cured inside an oven. 

\mbox{Fig.\,\ref{FIG02}(d)} shows the schematics of the two sides of the fabricated PCB. The top side of the PCB includes an MCU (Microchip Technology PIC16F18326-I/JQ) and an IMU (STMicroelectronics LSM6DSOXTR), which respectively perform onboard computation and sensing. We included programming ports in the circuit design, which are required to upload the operational \mbox{C-code} that runs the MCU; the programming is implemented using an MPLabs \mbox{PicKit\,$4$} \mbox{in-circuit} debugger tool and the \mbox{MPLabs~X~IDE~v$6.20$} developer environment. The bottom side of the PCB includes two MOSFETs, which function as switches that open and close the \mbox{current-flow} pathways to the SMA actuators when triggered by PWM signals generated by the MCU. As seen in \mbox{Fig.\,\ref{FIG02}(d)}, the bottom side of the PCB also includes a regulator that stabilizes the voltage delivered by the \mbox{Li-Ion} battery to power the MCU and IMU. The two sides of the PCB are connected through cylindrical \textit{vias} filled with conductive silver epoxy. Also, decoupling capacitors and \mbox{pull-up/down} resistors were included in the PCB design to ensure proper circuit functionality. To complete the assembly procedure, all the components of the prototype are connected together and secured using \textit{cyanoacrylate} (CA) glue (\mbox{Loctite\,$414$}) and the swimmer's hull is sealed by lining the inside and outside of all the structure's seams with CA glue. The SMA wire that drives each actuator is made of nitinol with a nominal transition temperature of \mbox{$90\,^{\circ}\ts{C}$} and diameter of \mbox{$0.0381\,\ts{mm}$} (Dynalloy~$90\,^{\circ}\ts{C}$~HT). In preliminary experiments, we found that driving both actuators of the Swima simultaneously resulted in drastic reductions in battery lifespan and circuit functionality due to the high \mbox{($\sim 500$\,mA)} instantaneous current draws from the Li-Ion battery. Due to this observation, we programmed the Swima to operate its tails alternately. 

\section{Autonomous Swimming}
\label{SECTION03}

\begin{figure*}[t!]
    \centering
    \includegraphics[width=\textwidth]{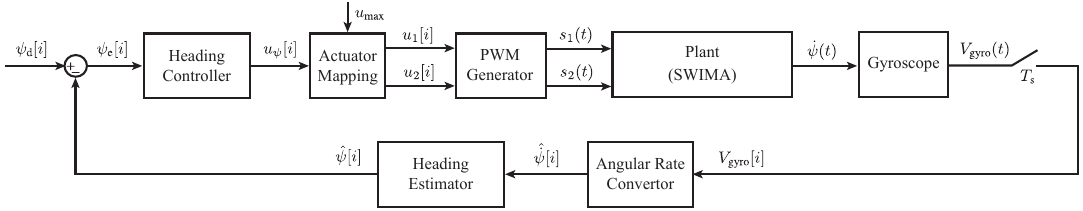}
    \caption{\textbf{Block diagram of the control scheme used in the \mbox{feedback-controlled} experiments.} In this scheme, the heading controller receives the heading error, $\psi_{\ts{e}}$, and computes the heading input signal, $u_{\psi}$. Then, the actuator mapping computes the DC values associated with each actuator, $u_1$ and $u_2$, which are used to generate the PWM analog signals, $s_1$ and $s_2$, that drive the \mbox{SMA-based} actuators. Depending on the heading angular velocity of the robot, $\dot{\psi}$, the gyroscope generates a voltage signal, $V_{\ts{gyro}}$, that is then sampled by the microcontroller and converted into a heading angular velocity estimate, $\hat{\dot{\psi}}$, through a static mapping provided by the IMU manufacturer. Last, we compute the heading estimate of the robot, $\hat{\psi}$, and use it to compute the heading error based on the desired heading, $\psi_{\ts{d}}$.}
    \label{FIG03}
    \vspace{-2ex}
\end{figure*}

In this section, we explain the control strategy used to perform the autonomous \mbox{feedback-controlled} swimming experiments, present and analyze the experimental swimming performance results, and discuss the endurance of the Swima through a \mbox{battery-duration} swimming test. For the \mbox{closed-loop} swimming experiments, we conducted heading-regulation and \mbox{$90$-degree} left- and right-turn maneuvers, which allowed us to determine the agility and performance of the robot measured in terms of average swimming speed, turning rate, and RMS value of the heading tracking error.

\subsection{Control Strategy}\label{SECTION03A}
To perform the \mbox{closed-loop} swimming experiments, we implemented the \mbox{feedback-control} scheme depicted in the block diagram shown in Fig.\,\ref{FIG03} where the objective is to track a desired heading signal, $\psi_{\ts{d}}$. Here, the output of the plant is the heading \mbox{angular-velocity} continuous signal, $\dot{\psi}(t)$, where $t$ denotes continuous time. This signal is then sensed by the gyroscope incorporated in the IMU, which generates an analog voltage, $V_{\ts{gyro}}$. This voltage signal is then sampled by the MCU and---using a static mapping provided by the IMU manufacturer---converted into an estimated heading angular velocity, $\hat{\dot{\psi}}[i]$, where $i$ is used to index discrete time. Next, the heading estimate of the robot is computed using a complementary filter using the current and previous estimates of the heading angular velocity,
\begin{align}
   \hat{\psi}[i] = \sum^{i}_{j=1}\left((1-c_{\ts{e}})\hat{\dot{\psi}}[j-1] + c_{\ts{e}}\hat{\dot{\psi}}[j]\right)\cdot T_{\ts{s}},
\end{align}
where $c_{\ts{e}}$ is a constant associated with the heading estimator; and $T_{\ts{s}}$ is the sampling time of the digital controller. We use this signal to then compute the heading error based on the desired heading, $\psi_{\ts{d}}$. Namely,
\begin{align}
    \psi_{\ts{e}}[i] = \psi_{\ts{d}}[i] - \hat{\psi}[i].
\end{align}
Next, the heading input signal is computed by a \textit{proportional-integral} (PI) control law according to
\begin{align}
    u_{\psi}[i] = c_{\ts{p}} \psi_{\ts{e}}[i] + c_{\ts{i}} \sum_{j=1}^{i} \psi_{\ts{e}}[j]\cdot T_{\ts{s}},
\end{align}
where $c_{\ts{p}}$ and $c_{\ts{i}}$ are positive scalar proportional and integral controller gains, respectively. The two control signals, $u_{\text{max}}$ and $u_\psi$---where $u_{\ts{max}}$ is the maximum duty-cycle value allowed to avoid burning the SMA wires---are operated by an actuator mapping that modulates the duty-cycle values associated with each actuator according to 
\begin{align}
    \begin{bmatrix}
        u_1\\
        u_2
    \end{bmatrix}=
    \begin{bmatrix}
        c_1 & c_1\\
        c_2 & -c_2
    \end{bmatrix}
    \begin{bmatrix}
        u_{\ts{max}}\\
        u_{\psi}
    \end{bmatrix},
\end{align}
where $c_1$ and $c_2$ are tuning constants that can be adjusted to compensate for variations in the resistance values of the two SMA-based actuators, ensuring that both soft tails visually have the same stroke envelope. The \mbox{duty-cycle} values $u_1$ and $u_2$ are then used to generate the analog PWM signals, $s_1(t)$ and $s_2(t)$, that excite the two actuators according to the chosen actuation frequency, $f_{\ts{a}}$. The PWM signals are generated using a custom code and sent through the digital output ports of the MCU to the SMA wires of each actuator. 

Although this control scheme does not directly control the position of the robot, in Section\,\ref{SECTION03B}, we show that by regulating the heading of the swimmer, we can compel the robot to follow a \mbox{straight-line} path, and that more complex maneuvers can be achieved by changing the desired heading in real time. Additionally, this control scheme can be evolved to allow position control by using the methods in~\cite{TrygstadCK2025}~and~\cite{BlankenshipEK2024}.

\begin{table}[t!]
\renewcommand{\arraystretch}{1.5}
\begin{center}
\caption{Control parameters used in the \mbox{closed-loop} experiments \label{Table01}}
\vspace{-2ex}
\centering
\begin{tabularx}{\linewidth}{Y|Y}
    \hline
    $c_{\ts{e}}$ (per unit) & $0.5$ \\
    \hline
    $u_{\ts{max}}$ (per unit) & $0.07$ \\
    \hline
    $c_{\ts{p}}$ (rad$^{-1}$) & $u_{\ts{max}}\cdot 0.2618^{-1}$ \\
    \hline
    $c_{\ts{i}}$ (rad$^{-1}\cdot$s$^{-1}$) & $0.05\cdot c_{\ts{p}}$ \\
    \hline
    $c_1$ (per unit) & $2$ \\
    \hline
    $c_2$ (per unit) & $1$ \\
    \hline
    $T_{\ts{s}}$ (s) & $0.001$ \\
    \hline
    $f_{\ts{a}}$ (Hz) & $2$ \\
    \hline
\end{tabularx}
\end{center}
\vspace{-4ex}
\end{table}

\begin{figure*}[t!]
    \centering
    \vspace{-1ex}
    \includegraphics[width=\textwidth]{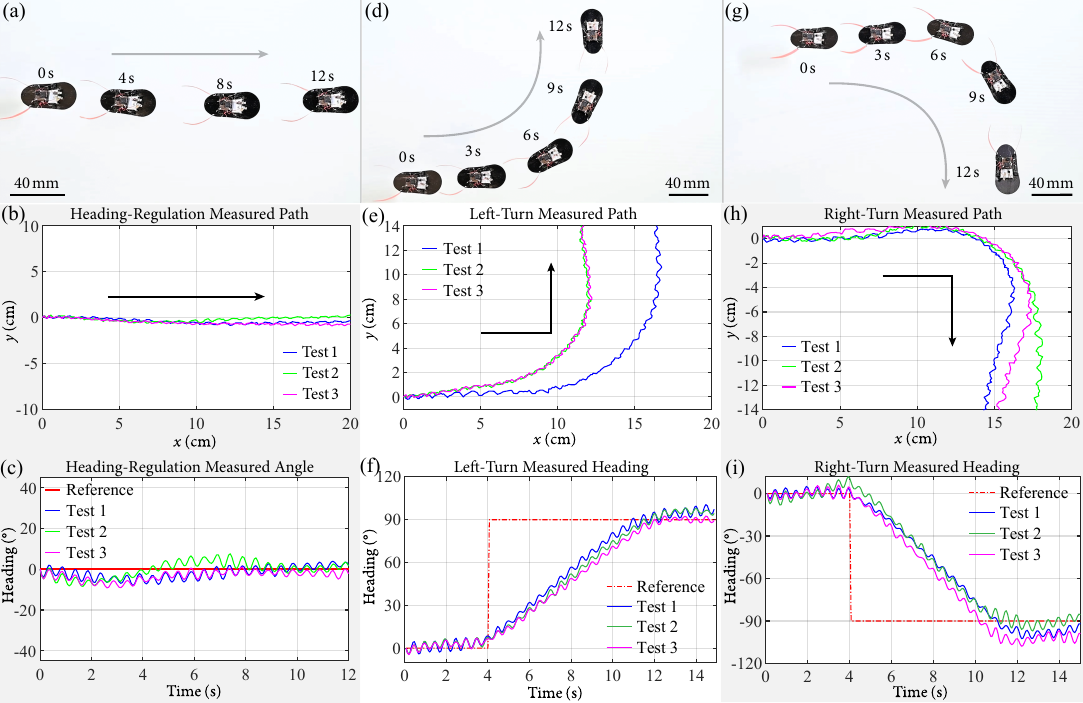}
    \caption{\textbf{Feedback-controlled swimming experiments with the Swima.} \textbf{(a)}~Composite image of frames taken at $4$-second intervals from video footage of the Swima following a straight-line path during Test\,1. \textbf{(b)}~Measured paths of three tests in which the swimmer was programmed to follow a \mbox{straight-line} path through \mbox{feedback-control} of its heading using the sensor information from the onboard IMU. \textbf{(c)}~Reference and measured heading of the three tests corresponding to (a). \textbf{(d)}~Composite image of frames taken at $4$-second intervals from video footage of the Swima following a \mbox{$90$-degree} \mbox{left-turn} maneuver. \textbf{(e)}~Measured path of an experiment where the swimmer was programmed to perform a \mbox{$90$-degree} \mbox{left-turn} maneuver through \mbox{feedback-control} of its heading. \textbf{(f)}~Reference and measured heading of the experiment corresponding to (c). \textbf{(g)} Composite image of frames taken at $4$-second intervals from video footage of the Swima following a \mbox{$90$-degree} \mbox{right-turn} maneuver. \textbf{(h)}~Measured path of an experiment where the swimmer was programmed to perform a \mbox{$90$-degree} \mbox{right-turn} maneuver through \mbox{feedback-control} of its heading. \textbf{(i)}~Reference and measured heading of the experiment corresponding to (e). Video footage of these experiments can be viewed in the supplementary movie available at \url{https://wsuamsl.com/resources/MovieSwima.mp4}.}
    \label{FIG04}
    \vspace{-0.5ex}
\end{figure*}

\subsection{Feedback-Controlled Swimming Experiments}\label{SECTION03B}
To execute the \mbox{feedback-controlled} swimming experiments with the Swima, we used the control parameters shown in Table\,\ref{Table01}. Fig.\,\ref{FIG04} summarizes the results of the Swima performing different maneuvers, including heading regulation and $90$-degree left and right turns. In specific, Fig.\,\ref{FIG04}(a) shows a composite image of frames taken at \mbox{$4$-s} intervals from video footage of Test\,1 of the Swima performing \mbox{heading-regulation} swimming using sensor information from its IMU and indirectly following a \mbox{straight-line} path; Fig.\,\ref{FIG04}(b) shows the measured paths of three tests of the same type, where repeatable performance can be observed; and Fig.\,\ref{FIG04}(c) shows the reference and measured heading of the three tests corresponding to Fig.\,\ref{FIG04}(b) where the robot was able to correct and follow a $0$-degree reference with an average RMS value of $6.5$\textdegree~in the three tests. Fig.\,\ref{FIG04}(d) shows a composite image of frames taken at $3$-s intervals from video footage of Test\,1 during which the Swima executes a \mbox{$90$-degree} \mbox{left-turn} maneuver; Fig.\,\ref{FIG04}(e) shows the measured path of the experiment corresponding to Fig.\,\ref{FIG04}(d), where a successful left turn can be observed; and Fig.\,\ref{FIG04}(f) shows the reference and measured heading of the test corresponding to the trajectories in Fig.\,\ref{FIG04}(e). In this case, we measured an average turning rate of \mbox{$11.3$\textdegree/s} across all three tests. Fig.\,\ref{FIG04}(g) shows a composite image of frames taken at $3$-s intervals from video footage of Test\,1 of the Swima executing a \mbox{$90$-degree} \mbox{right-turn} maneuver; Fig.\,\ref{FIG04}(h) shows the measured path of the experiment corresponding to Fig.\,\ref{FIG04}(g), where a successful right turn can be observed; and, Fig.\,\ref{FIG04}(i) shows the reference and measured heading of the tests corresponding to the trajectories in Fig.\,\ref{FIG04}(h). In this case, we measured an average turning rate of \mbox{$-14.0$\textdegree/s} across all three tests.

\subsection{Battery-Duration Swimming Test}\label{SECTION03C}

\begin{figure}[t!]
\vspace{0.75ex}
\begin{center}
\includegraphics[width=\linewidth]{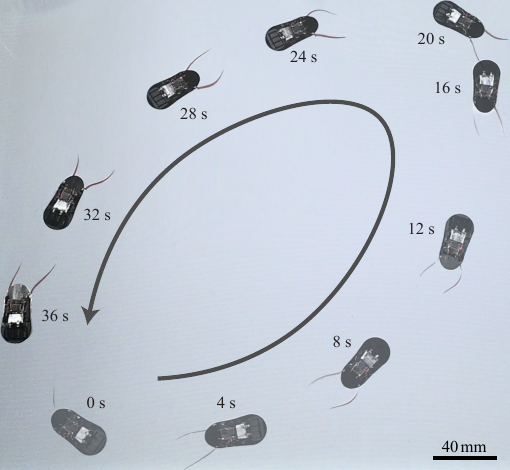}
\end{center}
\vspace{-1ex}
\caption{\textbf{Battery duration swimming experiment with the Swima.} The photographic composite of frames taken at $4$-second intervals represents the approximate swimming path covered by the Swima during one swimming lap during the $18$ minutes of continuous open-loop swimming at a frequency of $1$\,Hz. Video footage of this duration experiment can be viewed in the supplementary movie available at \url{https://wsuamsl.com/resources/MovieSwima.mp4}. \label{FIG05}}
\vspace{-2ex}
\end{figure}

To assess the battery lifespan of a Swima prototype during forward swimming, we programmed the robot to swim for an unlimited amount of time with alternating $1$-Hz PWM signals with a duty cycle of $5$\,\%. We fully charged the Li-ion battery to a voltage of $4.2$\,V and activated the custom-built PCB immediately before manually setting the robot in water using tweezers. We used an overhead camera (Nikon Z7II) to record the swimming experiment. To ensure the robot did not get trapped in a corner of the pool, we periodically adjusted the robot's trajectory by gently tapping its side with forceps to keep it swimming in circles around the pool. The experiment was run until the robot stopped functioning. Fig.\,\ref{FIG05} shows a photographic composite of frames taken at \mbox{$4$-s} intervals from overhead video of the Swima swimming a \emph{lap} in a pool in open loop. The battery voltage was read at the end of the experiment and was measured to be $3.7$\,V, indicating that a significant amount ($\sim50\,\%$) of energy was still remained unused. We hypothesize that this occurs because the high instantaneous current draw of the actuators transiently lowers the battery voltage below the drop-out threshold of the voltage regulator, causing the circuit to reset. Nonetheless, the Swima is capable of operating for up to $18$ minutes, which, considering the charge left in the Li-Ion battery, we estimate that the Swima consumed an average power on the order of $87$\,mW during this experiment. This agrees with the experimental measurements for the \mbox{SMA-based} actuators while operating in air \cite{LongwellCR2024}. Additionally, we estimate that during the swimming lap shown in Fig.\,\ref{FIG05}, the Swima maintained an average swimming speed on the order of \mbox{$22.4$\,mm/s}~\mbox{($0.56$\,Bl/s)}. \mbox{Sped-up} video footage of the entire swimming experiment can be viewed in the supplementary movie available at \url{https://wsuamsl.com/resources/MovieSwima.mp4}. 

\subsection{Discussion}

A comparison of all relevant autonomous insect-scale swimmers with lengths less than $100$\,mm \cite{SpinoP2024, TangL2026, LiuT2025, KimB2005, Xu2025Muscle} is shown in Table\,\ref{Table02}. We see that the Swima is the lightest among these swimmers and is the first subgram autonomous surface swimmer created to date. In terms of its average power consumption, that of the Swima is also among the lowest. Specifically, the only swimmer reported with a lower power consumption is the EEM robot~\cite{Xu2025Muscle}. 

The $18$-min single-charge operational time of the Swima is significant as it is among the longest operational time reported among state-of-the-art \emph{battery-powered} microrobots; for example, the \mbox{$2.7$-g} piezoelectric-based crawling platform in \cite{GoldbergB2018} is capable of only $4.5$ minutes of operation at its maximum locomotion speed while not being powered by external \mbox{radio-frequency-based} energy transfer, and the \mbox{$970$-mg} \mbox{DE-based} platform in \cite{JiX2019} is capable of $14$ minutes of continuous operation. Additionally, the $6.23$-gram swimmer in \cite{HartmanF2025} only operates for $10$ minutes on a single charge. The EEM swimming robot in \cite{Xu2025Muscle} is capable of operating at $1$\,Hz for up to $45\,\ts{min}$ on a single charge; however, it is important to note that while achieving this operational time, the EEM robot carries a much larger battery ($30$\,mAh), therefore, we predict that with this battery capacity, and assuming appropriate circuity is used to allow complete usage of the energy stored in the battery, the Swima would be capable of operating for up to $76\,\ts{min}$ with the battery in \cite{Xu2025Muscle}. However, the swimming performance of the EEM robot remains better than that of the Swima, suggesting there is room for improvement in the Swima propulsion system

While all the robots in Table\,\ref{Table02} demonstrate remarkable open-loop swimming performance and progress towards autonomous insect-scale swimmers, none of these robots have been demonstrated to be capable of autonomously following desired heading trajectories. The feedback-controlled swimming experiments presented in Section\,\ref{SECTION03A} are the first demonstrated to date for an autonomous insect-scale swimmer.

\begin{table}[t!]
\renewcommand{\arraystretch}{1.5}
\begin{center}
\caption{Comparison of relevant autonomous insect-scale swimmers\label{Table02}}
\vspace{-2ex}
\centering
\begin{tabularx}{\linewidth}{Y|c|c|c}
    \hline
    Robot  & Mass (mg) & Speed (Bl/s) & Power (mW) \\
    \hline
    Ball Robot\cite{SpinoP2024} & $1.2\times10^5$ & $3.9$ & $2.5\times10^3$
    \\
    \hline
    Leglessbot\cite{TangL2026} & $2.4\times10^5$ & $0.61$ & $1.9\times10^3$
    \\
    \hline
    \mbox{RobotPteropod\cite{LiuT2025}} & $3.4\times10^4$ & $1.2$ & $580$ 
    \\
    \hline
    \mbox{IPMC Tadpole\cite{KimB2005}} & $1.6\times10^4$ & $0.25$ & $1.2\times10^3$
    \\
    \hline
    Flat Swimmer\cite{HartmanF2025} & $6230$ & $1.14$ & $595$
    \\
    \hline
    EEM Robot\cite{Xu2025Muscle} & $2200$ & $1.37$ & $79$
    \\
    \hline
    This Work & $900$ & $0.56$ & $87$ 
    \\
    \hline
\end{tabularx}
\end{center}
\vspace{-2ex}
\end{table}

\section{Conclusion}
\label{SECTION04}

We presented a new \mbox{$900$-mg} surface swimmer, the Swima, which employs two \mbox{SMA-based} microactuators---in conjunction with a lightweight \mbox{custom-built} PCB that provides sensing, computation, and electronic capabilities---to locomote autonomously. A single charge of an onboard \mbox{Li-Ion} battery sustains about $18\,\ts{min}$ of untethered operation. Through swimming tests, we determined that the \mbox{Swima\hspace{-0.3ex}}~can swim at speeds of up to \mbox{$22.4$\,mm/s}~\mbox{($0.56$\,Bl/s)}, achieve turning rates of up to \mbox{$14$\textdegree/s}, and can follow $0$-degree heading reference trajectories with RMS values of tracking errors of about $6.5$\textdegree~across multiple tests. The Swima is (i) the first subgram microswimmer with onboard power, actuation, and computation developed to date, and (ii) the first insect-scale swimmer to demonstrate feedback-controlled heading tracking swimming experiments. Our future research aims to further improve the Swima's circuit to reduce instantaneous current draw on the Li-Ion battery and allow us to use the entirety of the stored charge, as well as improve the performance of the propulsion system.

\balance
\bibliographystyle{IEEEtran}
\bibliography{references}

@inproceedings{LongwellCR2024,
author="\mbox{C.~R.~Longwell} and \mbox{C.~K.~Trygstad} and \mbox{N.~O.~P\'erez-Arancibia}",
title="{\mbox{Power-efficient} actuation for \mbox{insect-scale} autonomous underwater Vehicles}",
booktitle="Proc. Int. Symp. Robot. Res. (ISRR)",
address="Long Beach, CA, USA",
month="Dec.",
year="2024",
note = "{Art.} no. 39"
}

@article{LiK2023,
author = "\mbox{K.\,Li} and \mbox{X.\,Zhou} and \mbox{Y.\,Liu} and \mbox{J.\,Sun} and \mbox{X.\,Tian} and 
\mbox{H.\,Zheng} and \mbox{L.\,Zhang} and \mbox{J.\,Deng} and \mbox{J.\,Liu} and \mbox{W.\,Chen} and \mbox{J.\,Zhao}",
title = "{A \mbox{$5$~cm-Scale} Piezoelectric Jetting Agile Underwater Robot}",
journal = "Adv. Intell. Syst.",
volume = "5",
number = "4",
note = "{Art.} no. 2200262",
month = "Apr.",
year = "2023"
}

@article{ChenY2017,
author = "\mbox{Y.\,Chen} and \mbox{H.\,Wang} and \mbox{E.\,F.\,Helbling} and \mbox{N.\,T.\,Jafferis} and \mbox{R.\,Zufferey} and {A.\,Ong} and {K.\,Ma} and {N.\,Gravish} and \mbox{P.\,Chirarattananon} and \mbox{M.\,Kovac} and \mbox{R.\,J.\,Wood}",
title = "{A Biologically Inspired, Flapping-Wing, Hybrid Aerial-Aquatic Microrobot}",
journal = "Sci. Robot.",
volume = "2",
number = "11",
note = "{Art.} no. eaao5619",
month ="Oct.",
year = "2017"
}

@inproceedings{TrygstadCK2023,
author = "\mbox{C.\,K.\,Trygstad} and \mbox{X.-T.\,Nguyen} and {N.\,O.\,P\'erez-Arancibia}",
title="{A New \mbox{$1$-mg} Fast Unimorph \mbox{SMA-Based} Actuator for Microrobotics}", 
booktitle = "Proc. IEEE/RSJ Int. Conf. Intell. Robots Syst. (IROS)", 
address = "Detroit, MI, USA",
month = "Oct.",
year = "2023",
pages = "2693--2700"
}

@inproceedings{BlankenshipEK2024,
author = "\mbox{E.~K.~Blankenship} and \mbox{C.~K.~Trygstad} and \mbox{F.~M.~F.~R.\,Gon\c{c}alves} and \mbox{N.~O.~P\'erez-}~\mbox{Arancibia}",
title = "{VLEIBot: A New \mbox{$45$-mg} Swimming Microrobot Driven by a Bioinspired Anguilliform Propulsor}",
booktitle = "Proc. IEEE Int. Conf. Robot. Autom. (ICRA)",
address = "Yokohama, Japan",
month = "May",
year = "2024", 
pages = "6014--6021"
}

@inproceedings{TrygstadCK2024,
author = "\mbox{C.\,K.\,Trygstad} and \mbox{E.\,K.\,Blankenship} and \mbox{N.\,O.\,P\'erez-Arancibia}",
title = "{A New \mbox{$10$-mg} \mbox{SMA-Based} Fast Bimorph Actuator for Microrobotics}",
booktitle = "Proc. IEEE/RSJ Int. Conf. Intell. Robots Syst. (IROS)", 
address = "Abu Dhabi, UAE",
month = "Oct.",
year = "2024",
pages = "1349--1356"
}

@inproceedings{TrygstadCK2025,
author = "\mbox{C.\,K.\,Trygstad} and \mbox{C.\,R.\,Longwell} and \mbox{F.\,M.\,F.\,R.\,Gon\c{c}alves} and \mbox{E.\,K.\,Blankenship} and \mbox{N.\,O.\,P\'erez-Arancibia}",
title = "{Feedback Control of a Single-Tail Bioinspired 59-mg Swimmer}",
booktitle = "Proc. IEEE/RSJ Int. Conf. Intell. Robots Syst. (IROS)", 
address = "Hangzhou, China",
month = "Dec.",
year = "2025",
note = "To Appear"
}

@article{BenaRM2023,
author = "\mbox{R.~M.~Bena} and \mbox{X.~Yang} and \mbox{A.~A.~Calder\'on} and \mbox{N.~O.~P\'erez-Arancibia}", 
title = "{High-Performance \mbox{Six-DOF} Flight Control of the Bee\textsuperscript{++}: An Inclined-Stroke-Plane Approach}", 
journal = "IEEE Trans. Robot.",
volume = "39",
number = "2",
pages = "1668--1684",
month = "Apr.",
year = "2023"
}

@article{RenZ2022,
author = "\mbox{Z.\,Ren} and \mbox{S.\,Kim} and \mbox{X.\,Ji} and \mbox{W.\,Zhu} and \mbox{F.\,Niroui} and \mbox{J.\,Kong} and \mbox{Y.\,Chen}",
title = "{A High-Lift Micro-Aerial-Robot Powered by Low-Voltage and Long-Endurance Dielectric Elastomer Actuators}",
journal = "Adv. Mat.",
volume = "34",
number = "7",
note = "{Art.} no. 2106757",
month = "Feb.",
year = "2022"
}

@article{WuY2019,
author = "\mbox{Y.\,Wu} and \mbox{J.\,K.\,Yim} and \mbox{J.\,Liang} and \mbox{Z.\,Shao} and \mbox{M.\,Qi} and \mbox{J.\,Zhong} and \mbox{Z.\,Luo} and \mbox{X.\,Yan} and \mbox{M.\,Zhang} and \mbox{X.\,Wang} and \mbox{R.\,S.\,Fearing} and \mbox{R.\,J.\,Full} and \mbox{L.\,Lin}",
title = "{\mbox{Insect-Scale} Fast Moving and Ultrarobust Soft Robot}",
journal = "Sci. Robot.",
volume = "4",
number = "32",
note = "{Art.} no. eaax1594",
month = "Jul.",
year = "2019"
}

@inproceedings{ZhouW2020,
author = "\mbox{W.\,Zhou} and \mbox{N.\,Gravish}",
title = "{Soft Microrobotic Transmissions Enable Rapid Ground-Based Locomotion}",
booktitle = "Proc. IEEE/RSJ Int. Conf. Intell. Robots Syst. (IROS)",
address = "Las Vegas, NV, USA",
Month = "Oct.",
year = "2020",
pages = "7874--7880"
}

@article{BenaRM2021,
author="\mbox{R.\,M.\,Bena} and \mbox{X.-T.\,Nguyen} and {A.\,Rigo} and \mbox{A.\,A.\,Calder\'on} and \mbox{N.\,O.~P\'erez-Arancibia}",
title="{SMARTI: A \mbox{$60$-mg} Steerable Robot Driven by High-Frequency Shape-Memory Alloy Actuation}", 
journal = "IEEE Robot. Automat. Lett.", 
volume = "6",
number = "4",
pages = "8173--8180",
month = "Oct.",
year = "2021"
}

@article{YangX2020,
author = "\mbox{X.\,Yang} and \mbox{L.\,Chang} and \mbox{N.\,O.\,\mbox{P\'erez-Arancibia}}",
title = "{An 88-Milligram Insect-Scale Autonomous Crawling Robot Driven by a Catalytic Artificial Muscle}",
journal = "Sci. Robot.",
volume = "5",
number = "45",
note = "{Art.} no. eaba0015",
month = "Aug.",
year = "2020"
}

@inproceedings{XuT2013,
author = "\mbox{T.\,Xu} and \mbox{G.\,Hwang} and \mbox{N.\,Andreff} and \mbox{S.\,R\'egnier}",
title = "{The Rotational Propulsion Characteristics of Scaled-Up Helical Microswimmers With Different Heads and Magnetic Positioning}", 
booktitle = "Proc. IEEE/ASME Int. Conf. Adv. Intell. Mechatron. (AIM)", 
address = "Wollongong, NSW, Australia",
month = "Jul.",
year = "2013",
pages = "1114--1120",
}

@article{ZhangL2010,
author = "\mbox{L.~Zhang} and \mbox{T.~Petit} and \mbox{Y.~Lu} and \mbox{B.~E.~Kratochvil} and \mbox{K.~E.~Peyer} and \mbox{R.~Pei} and \mbox{J.~Lou} and \mbox{B.~J.~Nelson}",
title = "{Controlled Propulsion and Cargo Transport of Rotating Nickle Nanowires Near a Patterned Solid Surface}",
journal = "ACS Nano",
volume = "4",
number = "10",
pages = "6228--6234",
month = "Sep.",
year = "2010"
}

@article{LiuW2010,
author = "\mbox{W.\,Liu} and \mbox{X.\,Jia} and \mbox{F.\,Wang} and \mbox{Z.\,Jia}",
title = "{An In-Pipe Wireless Swimming Microrobot Driven by Giant Magnetostrictive Film}",
journal = "Sens. Actuators A: Phys.",
volume = "160",
number = "2",
pages = "101--108",
month = "May",
year = "2010"
}

@article{GoldbergB2018,
author = "\mbox{B.\,Goldberg} and \mbox{R.\,Zufferey} and \mbox{N.\,Doshi} and \mbox{E.\,F.\,Helbling} and \mbox{G.\,Whittredge} and \mbox{M.\,Kovac} and \mbox{R.\,J.\,Wood}",
title = "{Power and Control Autonomy for High-Speed Locomotion With an Insect-Scale Legged Robot}",
journal = "IEEE Robot. Automat. Lett.",
volume = "3",
number = "2",
pages = "987--993",
month = "Apr.",
year = "2018"
}

@inproceedings{JohnsonK2023,
author="\mbox{K.\,Johnson} and \mbox{Z.\,Englehardt} and \mbox{V.\,Arroyos} and \mbox{D.\,Yin} and \mbox{S.\,Patel} and \mbox{V.\,Iyer}",
title = "{MilliMobile: An Autonomous Battery-Free Wireless Microrobot}",
booktitle = "Proc. 29th Annu. Int. Conf. Mob. Comput. Netw. (MOBICOM) ",
address = "Madrid, Spain",
month = "Oct.",
year = "2023",
pages = "1360--1375",
}

@article{JiX2019,
author = "\mbox{X.~Ji} and \mbox{X.~Liu} and \mbox{V.~Cacucciolo} and \mbox{M.~Imboden} and \mbox{Y.~Civet} and \mbox{A.~E.~Haitami} and \mbox{S.~Cantin} and \mbox{Y.~Perriard} and \mbox{H.~Shea}",
title = "{An Autonomous Untethered Fast Soft Robotic Insect Driven by Low-Voltage Dielectric Elastomer Actuators}",
journal = "Sci. Robot.",
volume = "4",
number = "37",
note = "{Art.} no. eaaz6451",
month = "Dec.",
year = "2019"
}

@inproceedings{SpinoP2024,
author = "\mbox{P.\,Spino} and \mbox{D.\,Rus}", 
title = "{Towards Centimeter-Scale Underwater Mobile Robots: An Architecture for Capable {\textmu}AUVs}",
booktitle = {Proc. IEEE Int. Conf. Robot. Autom. (ICRA)},
address = "Yokohama, Japan",
month = "May",
year = "2024",
pages = "1484--1490",
}

@article{BerlingerF20210-Sci,
author = "\mbox{F.\,Berlinger} and \mbox{M.\,Gauci} and \mbox{R.\,Nagpal}",
title = "{Implicit Coordination for $3$D Underwater Collective Behaviors in a Fish-Inspired Robot Swarm}",
journal = "Sci. Robot.",
volume = "6",
number = "50",
note = "{Art.} no. eabd8668",
month = "Jan.",
year = "2021",
}

@article{BerlingerF2021-Bio,
author = "\mbox{F.\,Berlinger} and \mbox{M.\,Saadat} and \mbox{H.\,Haj-Hariri} and \mbox{G.\,V.\,Lauder} and \mbox{R.\,Nagpal}",
title = "{Fish-Like Three-Dimensional Swimming With an Autonomous, Multi-Fin, and Biomimetic Robot}",
journal = "Bioinspir. Biomim.",
volume = "16",
number = "2",
note = "{Art.} no. 026018",
month = "Mar.",
year = "2021"
}

@inproceedings{LiZ2024,
author = "\mbox{Z.\,Li} and \mbox{Y.\,Zhang} and \mbox{S.\,He} and \mbox{X.\,Zhu} and \mbox{T.\,Wei} and \mbox{C.\,Hu}",
title = "{Centimeter-Scale Submarine Robot for Monitoring Coral Reef Ecosystem}", 
booktitle = "Proc. IEEE Int. Conf. Mechatron. Autom. (ICMA)",
address = "Tianjin, China",
month = "Aug.",
year = "2024",
pages = "345--350"
}

@article{GiorgioFS2013,
author = "\mbox{F.\,Giorgio-Serchi} and \mbox{A.\,Arienti} and \mbox{C.\,Laschi}",
title = "{Biomimetic Vortex Propulsion: Toward the New Paradigm of Soft Unmanned Underwater Vehicles}", 
journal = "IEEE/ASME Trans. Mechatron.", 
volume = "18",
number = "2",
pages = "484--493",
month = "Apr.",
year = "2013"
}

@article{CenL2013,
author = "\mbox{L.\,Cen} and \mbox{A.\,Erturk}",
journal = "{Bioinsp. Biomim.}", 
title = "{Bio-Inspired Aquatic Robotics by Untethered Piezohydroelastic Actuation}", 
volume = "8",
number = "1",
note = "{Art.} no. 016006",
month = "Mar.",
year = "2013"
}

@article{WangT2023,
author = "\mbox{T.\,Wang} and \mbox{H.\,Joo} and \mbox{S.\,Song} and \mbox{W.\,Hu} and \mbox{C.\,Keplinger} and \mbox{M.\,Sitti}",
title = "{A Versatile Jellyfish-Like Robotic Platform for Effective Underwater Propulsion and Manipulation}",
journal = "Sci. Adv.",
volume = "9",
number = "15",
pages = "{Art.} no. eadg0292",
month = "Apr.",
year = "2023"
}

@article{VillanuevaA2011,
author = "\mbox{A.\,Villanueva} and \mbox{C.\,Smith} and \mbox{S.\,Priya}",
journal = "{Bioinsp. Biomim.}", 
title = "{A biomimetic robotic jellyfish (Robotjelly) actuated by shape memory alloy composite actuators}", 
volume = "6",
number = "3",
note = "{Art.} no. 036004",
month = "Sep.",
year = "2011"
}

@article{LiuT2025,
author = "\mbox{T.\,Liu} and \mbox{Y.\,Liu} and \mbox{R.\,Zeng} and \mbox{B.\,Gan} and \mbox{M.\,Zheng} and \mbox{H.\,Li} and \mbox{S.\,Qu} and \mbox{H.\,Zhou}",
title = "{A bioinspired multimotion modality underwater microrobot}",
journal = "Sci. Adv.",
volume = "11",
number = "19",
note = "{Art.} no. eadu2527",
month = "May",
year = "2025"
}

@inproceedings{TangY2017,
author = "\mbox{Y.\,Tang} and \mbox{L.\,Qin} and \mbox{X.\,Li} and \mbox{C.-M.\,Chew} and \mbox{J.\,Zhu}",
title = "{A Frog-inspired Swimming Robot Based on Dielectric Elastomer Actuators}", 
booktitle = "Proc. IEEE/RSJ Int. Conf. Intell. Robots Syst. (IROS)",
address = "Vancouver, BC, Canada",
month = "Sep.",
year = "2017",
pages = "2403--2408"
}

@article{ZiolkowskiA1993,
title = "{Theoretical analysis of efficiency of shape memory alloy heat engines (based on constitutive models of pseudoelasticity)}",
author = "\mbox{A.~Zi{\'o}{\l}kowski}",
journal = "\mbox{Mech.~Mater.}",
volume = "16",
year = "1993",
pages = "365--377"
}

@article{JaniJM2014,
title = "{A review of shape memory alloy research, applications and opportunities}",
author = "\mbox{J.~M.~Jani} and \mbox{M.~Leary} and \mbox{A.~Subic} and \mbox{A.~M.~Gibson}",
journal = "\mbox{Mater.~Des.}",
volume = "56",
year = "2014",
pages = "1078--1113"
}

@article{KarpelsonM2012,
title = "{Driving high voltage pie-zoelectric actuators in microrobotic applications}", 
author = "\mbox{M.~Karpelson} and \mbox{G.-Y.~Wei} and \mbox{R.~J.~Wood}",
journal = "{Sens. Actuators A Phys.}", 
volume = "176",
year = "2012",
pages = "78--89"
}

@inproceedings{RenZ2023,
title = "{A lightweight \mbox{high-voltage} boost circuit for soft-actuated \mbox{micro-aerial-robots}}", 
author = "\mbox{Z.~Ren} and \mbox{J.~Yang} and \mbox{S.~Kim} and \mbox{Y.-H.~Hsiao} and \mbox{J.~Lang} and \mbox{Y.~Chen}",
booktitle = "{Proc. IEEE Int. Conf. Robot. Autom. (ICRA)}",
address = "{London, UK}",
year = "2023",
pages = "3397--3403"
}

@article{HubbardJJ2014,
title = "{Monolithic IPMC fins for propulsion and ma-neuvering in bioinspired underwater robotics}", 
author = "\mbox{Hubbard,~J.~J.} and \mbox{Fleming,~M.} and \mbox{Palmre,~V.} and \mbox{Pugal,~D.} and \mbox{Kim,~K.~J.} and \mbox{Leang,~K.~K.}",
journal = "{IEEE J. Oceanic Eng.}", 
volume = "39",
year = "2014",
pages = "540--551"
}

@article{GuoS2003,
title = "{A new type of \mbox{fish-like} underwater microrobot}", 
author = "\mbox{Guo,~S.} and \mbox{Fukuda,~T.} and \mbox{Asaka,~K.}",
journal = "{IEEE Trans. Mechatron.}", 
volume = "8",
year = "2003",
pages = "136--141"
}

@article{HeQ2011,
title = "{Experimental study and model analysis of the performance of IPMC membranes with various thickness}",
author = "\mbox{He,~Q.} and \mbox{Yu,~M.} and \mbox{Song,~L.} and \mbox{Ding,~H.} and \mbox{Zhang,~X.} and \mbox{Dai,~Z.}",
journal = "{J. Bionic Eng.}",
volume = "8",
year = "2011",
pages = "77--85"
}

@article{KimB2005,
title = "{A biomimetic undulatory tadpole robot using ionic \mbox{polymer–metal} composite actuators}",
author = "\mbox{Kim,~B.} and \mbox{Kim,~D.-H.} and \mbox{Jung,~J.} and \mbox{Park,~J.-O.}",
journal = "{Smart Mater. Struct.}",
volume = "14",
year = "2005",
pages = "1579--1585"
}

@article{ChenZ2012,
title = "{\mbox{Bio-inspired} robotic manta ray powered by ionic \mbox{polymer–metal} composite artificial muscles}",
author = "\mbox{Chen,~Z.} and \mbox{Um,~T.~I.} and \mbox{Bart-Smith,~H.}",
journal = "{Int. J. Smart Nano Mater.}",
volume = "3",
year = "2012",
pages = "296--308"
}

@article{ChenA2010,
title = "{Modeling of biomimetic robotic fish propelled by an ionic \mbox{polymer–metal} composite caudal fin}", 
author = "\mbox{Chen,~Z.} and \mbox{Shatara,~S.} and \mbox{Tan,~X.}",
journal = "{IEEE/ASME Trans. Mechatron.}", 
volume = "15",
year = "2010",
pages = "448--459"
}

@article{ChenZ2011,
title = "{A novel fabrication of ionic \mbox{polymer–metal} composite membrane actuator capable of \mbox{3-dimensional} kinematic motions}", 
author = "\mbox{Chen,~Z.} and \mbox{Um,~T.~I.} and \mbox{Bart-Smith,~H.}",
journal = "{Sens. Actuators A}", 
volume = "168",
year = "2011",
pages = "131--139"
}

@inproceedings{TrygstadCK2025ICAR,
author = "\mbox{C.\,K.\,Trygstad} and {N.\,O.\,P\'erez-Arancibia}",
title = "{Force Characterization of Insect-Scale Aquatic Propulsion Based on Fluid-Structure Interaction}",
booktitle = "Proc. IEEE Int. Conf. Adv. Robot. (ICAR)", 
address = "San Juan, Argentina",
month = "Dec.",
year = "2025",
pages = "764--771"
}

@article{ChitikenaH2023,
title = "{Robotics in search and rescue (SAR) operations: an ethical and design perspective framework for response phase}",
author = "\mbox{Chitikena,~H.} and \mbox{Sanfilippo,~F.} and \mbox{Ma,~S.}",
journal = "{Appl. Sci.}",
volume = "13",
year = "2023",
note = "{Art. no. 1800}"
}

@article{AitkenJM2021,
title = "{Simultaneous localization and mapping for inspection robots in water and sewer pipe networks: a review}",
author = "\mbox{Aitken,~J.~M.} and \mbox{Evans,~M.~H.} and \mbox{Worley,~R.} and \mbox{Edwards,~S.} and \mbox{Zhang,~R.} and \mbox{Dodd,~T.} and \mbox{Mihaylova,~L.} and \mbox{Anderson,~S.~R.}",  
journal = "{IEEE Access}", 
volume = "9",
year = "2021",
pages = "140173--140198"
}

@article{UrsoM2023,
title = "{Smart micro- and nanorobots for water purification}",
author = "\mbox{Urso,~M.} and \mbox{Ussia,~M.} and \mbox{Pumera,~M.}",
journal = "{Nat. Rev. Bioeng.}",
volume = "1",
year = "2023",
pages = "236--251"
}

@article{Liu2024BHMBot,
    title = "{A wireless controlled robotic insect with ultrafast untethered running speeds}",
    author = {Zhiwei Liu and Wencheng Zhan and Xinyi Liu, Yangsheng Zhu and Mingjing Qi and Jiaming Leng and Lizhao Wei and Shousheng Han and Xiaoming Wu and Xiaojun Yan},
    journal = "{nat. comm.}",
    volume = "15",
    year = "2024",
    note = "{Art. no. 3815}"
}

@article{Xu2025Muscle,
author="{Xu,~C.} and {Cao,~Y.} and {Zhao,~J.} and {Cao,~Y.} and {Huang,~Y.} and {Lin,~Y.} and {Wang,~D.} and {Zhang,~Z.} and {Jiang,~H.}",
title="{Muscle-inspired elasto-electromagnetic mechanism in autonomous insect robots}",
journal="Nature Comm.",
volume="16",
number="1",
note="{Art.} no. 6813",
month = "Jul.",
year="2025"
}

@article{TangL2026,
author = {Tang,~L. and Yang,~Y. and Li,~B. and Zhang,~B. and He,~Q. and Ren,~H. and Li,~Y.},
title = {Inertia-driven amphibious robot with asymmetric microundulatory fin arrays},
journal = {Science Advances},
volume = {12},
number = {8},
pages = {eaea2222},
year = {2026},
}

@article{HartmanF2025,
author = {Florian Hartmann  and Mrudhula Baskaran  and Gaetan Raynaud  and Mehdi Benbedda  and Karen Mulleners  and Herbert Shea},
title = {Highly agile flat swimming robot},
journal = {Science Robotics},
volume = {10},
number = {99},
pages = {eadr0721},
year = {2025},
}

\end{document}